\documentclass[11pt]{article}

\usepackage[final]{acl}

\usepackage{times}
\usepackage{latexsym}

\usepackage[T1]{fontenc}
\usepackage[utf8]{inputenc}

\usepackage{microtype}

\usepackage{inconsolata}

\usepackage{graphicx}

\usepackage{amsmath,amssymb,amsfonts}
\usepackage{algorithmic}
\usepackage{arydshln}
\usepackage{makecell}
\usepackage{multirow}
\usepackage{tcolorbox}
\tcbuselibrary{breakable}
\usepackage{pifont}
\usepackage{xcolor}
\definecolor{darkgreen}{rgb}{0.0, 0.5, 0.0}

\title{LitE-SQL: A Lightweight and Efficient Text-to-SQL Framework with Vector-based Schema Linking and Execution-Guided Self-Correction}

\author{Shengmin Piao\textsuperscript{*}, Jieun Lee\textsuperscript{*}, Sanghyun Park\textsuperscript{\textdagger} \\
    Yonsei University \\
    Seoul, South Korea \\
    \texttt{\{shengminp, jieun199624, sanghyun\}@yonsei.ac.kr} \\}

\begin{document}
\maketitle
\renewcommand{\thefootnote}{}
\footnotetext{\textsuperscript{*} Same contribution.}
\footnotetext{\textsuperscript{\textdagger} Corresponding author.}
\renewcommand{\thefootnote}{\arabic{footnote}}
\begin{abstract}
The Text-to-SQL task translates natural language questions into SQL queries, enabling intuitive database interaction for non-experts. While recent methods leveraging Large Language Models (LLMs) achieve strong performance, their reliance on proprietary models raises concerns about deployment feasibility and data privacy. In this work, we introduce LitE-SQL, a Lightweight and Efficient framework with two components: %(i) a Schema Retriever that performs efficient schema linking using a vector database of pre-computed schema embeddings, and (ii) a SQL Generator fine-tuned in two stages—supervised fine-tuning followed by execution-guided reinforcement—enabling self-correction without costly multi-candidate generation.
(i) a Schema Retriever that performs efficient schema linking using a vector database of pre-computed schema embeddings, optimized with a hard-negative supervised contrastive objective to distinguish semantically similar but functionally irrelevant columns, and (ii) a SQL Generator fine-tuned in two stages—supervised fine-tuning followed by execution-guided reinforcement—enabling execution-guided self-correction without multi-candidate sampling, which is commonly required by prior LLM-based approaches.
On BIRD, LitE-SQL achieves 72.10\% execution accuracy, and on Spider 1.0 it reaches 88.45\%, demonstrating comparable or superior performance to LLM-based methods despite using 2× to 30× fewer parameters. Our findings demonstrate that high-quality Text-to-SQL generation is feasible with lightweight models, offering a practical solution for privacy-sensitive and resource-constrained settings. Code is available at \href{https://github.com/shengminp/LitE-SQL}{https://github.com/shengminp/LitE-SQL}.
\end{abstract}

\section{Introduction}

Relational databases remain a cornerstone of modern data management, with Structured Query Language (SQL) serving as the standard interface for querying structured data. However, composing SQL queries poses a barrier for non-expert users. This has motivated growing interest in Text-to-SQL, which translates natural language questions into SQL queries, enabling intuitive interaction with relational databases \cite{hong2024next,li2024codes}. 

Early approaches relied on rule-based systems~\cite{li2014constructing}, whereas advances in deep neural networks substantially improved semantic parsing and SQL generation~\cite{li2023can}. Despite these gains, existing methods still struggle with cross-domain generalization and complex query generation~\cite{singh2025survey}. Recently, Large Language Models (LLMs) have demonstrated remarkable performance via in-context learning capabilities~\cite{brown2020language}. These approaches enhance SQL generation by designing structured prompts and incorporating database schema information into the input context~\cite{li2023can, pourreza2023din, 10.14778/3641204.3641221, 10.14778/3681954.3682003, lee-etal-2025-mcs, DBLP:journals/corr/abs-2405-16755, maamari2024the, pourreza2025chasesql}.

Although LLM-based methods mitigate several limitations of earlier approaches, they introduce new concerns—particularly data privacy and deployment constraints arising from reliance on proprietary models~\cite{hong2024next} such as ChatGPT \cite{achiam2023gpt} or Gemini \cite{team2023gemini}. This reliance risks information leakage in domains with strict data security requirements. Developing custom LLMs is a possible remedy but incurs substantial development and maintenance costs. These limitations underscore the need for lightweight and efficient alternatives.

To this end, we identify two core technical challenges: \textbf{(i) Contextual Expansion of Schema Information.} Incorporating complete schema information into model inputs is critical for accurate generation but substantially increases input length and, due to the quadratic complexity of the self-attention mechanism \cite{vaswani2017attention}, results in high memory consumption. Such long-context inputs impose a considerable computational burden on lightweight models, hindering efficient processing in resource-constrained environments. \textbf{(ii) Computational Footprint of Large-Parameter LLMs.} While LLM-based methods achieve strong performance in SQL generation, each inference incurs significant computational cost, which is often exacerbated by strategies such as multi-candidate generation. These limitations highlight the need for lightweight alternatives that offer competitive performance while remaining feasible in constrained computational settings.

To address these challenges, we propose \textbf{LitE-SQL}, a \textbf{Li}ghtweigh\textbf{t} and \textbf{E}fficient framework comprising a \textit{Schema Retriever} and a \textit{SQL Generator}. We design a retrieval-based schema linking approach using a vector database. Unlike existing methods that rely on LLMs~\cite{pourreza2025chasesql,DBLP:journals/corr/abs-2405-16755} or heavyweight encoders~\cite{li2024codes}, our Schema Retriever stores pre-computed schema embeddings in a vector database and, at inference, encodes only the question to retrieve relevant columns through semantic similarity search. To our knowledge, this is the first work to fully exploit a vector-database-driven retriever for schema linking in Text-to-SQL. For the SQL Generator, we develop a two-stage fine-tuning strategy that tackles computational cost by employing lightweight language models. In the first stage, the model generates SQL conditioned on the question and retrieved schema; in the second stage, it refines the outputs using execution feedback. This execution-guided self-correction enables efficient error correction without relying on multi-candidate generation.

We conduct comprehensive experiments on two benchmark datasets, BIRD \cite{li2023can} and Spider 1.0 \cite{yu-etal-2018-spider}, to evaluate the effectiveness of LitE-SQL. On BIRD, LitE-SQL achieves an execution accuracy of 72.10\%, and on Spider 1.0, it reaches 88.45\%. These results are competitive with LLM-based approaches while using approximately 2× to 30× fewer parameters, highlighting its efficiency and suitability for resource-constrained environments. Furthermore, the Schema Retriever achieves retrieval quality on par with or superior to existing schema linking approaches, and the SQL Generator consistently corrects erroneous SQL queries using execution feedback, as evidenced by a measurable reduction in error rates in self-correction analyses. These findings indicate that LitE-SQL offers a practical and effective lightweight framework for Text-to-SQL.

% The key contributions of this work are as follows:
% \begin{itemize}
%     \item We present a vector-database-driven Schema Retriever that identifies relevant schema information via semantic similarity, representing (to our knowledge) the first such application to schema linking in Text-to-SQL.
%     \item We develop a two-stage fine-tuning strategy for the SQL Generator, enabling lightweight models to self-correct with execution feedback without multi-candidate generation.
% \end{itemize}

The key contributions of this work are as follows:
\begin{itemize}
    \item We identify core challenges in deploying lightweight models for Text-to-SQL and propose LitE-SQL, a lightweight and efficient framework consisting of a Schema Retriever and a SQL Generator.
    \item We introduce a hard-negative supervised contrastive learning objective tailored for schema-level retrieval in Text-to-SQL.
    \item We develop a two-stage fine-tuning strategy for the SQL Generator, enabling lightweight models to self-correct with execution feedback without multi-candidate sampling.
\end{itemize}

\section{Related Work}
\subsection{Schema Linking} \label{subsec:related_schema_linking}
Schema linking identifies relevant schema elements for a given question, a key step toward generating correct SQL. Early methods trained classifiers to rank schema elements~\cite{li2024codes,pourreza-rafiei-2024-dts}, relying on supervised filters to score column and table relevance.

Recent approaches leverage LLMs through in-context learning, injecting full or partial schema information into prompts~\cite{10.14778/3641204.3641221, li2025omnisql, wang-etal-2025-mac}. Some adopt multi-step prompting—such as generating dummy SQL or asking models to justify schema selections—to identify relevant columns~\cite{qu-etal-2024-generation,lee-etal-2025-mcs}. CHESS~\cite{DBLP:journals/corr/abs-2405-16755} introduces a multi-agent framework where an Information Retriever collects context and a Schema Selector identifies relevant schema elements. Distillery~\cite{maamari2024the} questions the need for explicit schema linking, showing that strong LLMs can infer schema implicitly. However, smaller models degrade significantly without filtering irrelevant columns, underscoring the importance of schema selection for lightweight models.

Despite their effectiveness, these methods incur substantial overhead by re-encoding the schema for every input. In addition, CHESS and CHASE-SQL employ embedding similarity via vector databases, but mainly for keyword retrieval, leaving final schema selection to LLMs. In contrast, we leverage vector databases as the core retriever for schema linking, explicitly designed for lightweight Text-to-SQL models, while encoding only the input question at inference. To further exploit vector databases effectively, we optimize the embedding space with a tailored objective, improving retrieval quality and overall Text-to-SQL performance.

\subsection{SQL Generation} \label{subsec:related_sql_generation}
SQL generation methods with language models fall broadly into in-context learning and fine-tuning.

\textbf{In-context learning.} LLMs demonstrate strong SQL generation capabilities when guided by carefully designed prompts. A basic strategy provides the database schema alongside an instruction to generate the query \cite{li2023can}. Few-shot prompting further improves performance by adding few examples \cite{pourreza2023din, maamari2024the, 10.14778/3681954.3682003}. To handle complex questions, structured prompts decompose questions into sub-questions \cite{pourreza2023din, wang-etal-2025-mac, pourreza2025chasesql}, or employ chain-of-thought prompting \cite{wei2022chain}, which generates SQL queries incrementally\cite{li2023can, wang-etal-2025-mac, pourreza2025chasesql}. Many methods further apply self-consistency \cite{wang2022self}, generating multiple candidates and selecting the majority. While effective, these strategies are computationally expensive due to multiple LLM calls per question. In contrast, our work explores fine-tuned lightweight models that avoid such overhead while preserving competitive accuracy.

\textbf{Fine-tuning.} Fine-tuning pre-trained models provides stronger domain adaptation, fine-grained control, and better suitability for privacy-sensitive deployments. Given that powerful models such as GPT-4 \cite{achiam2023gpt} remain proprietary, recent efforts emphasize open-source models fine-tuned for SQL generation \cite{li2024codes, li2025omnisql, pourreza-rafiei-2024-dts, pourreza2025reasoning}. To enhance the understanding between natural language and SQL, fine-tuning with a language modeling objective on SQL data offers a direct and effective way. Building on this foundation, methods such as data augmentation \cite{li2025omnisql}, incremental pretraining \cite{li2024codes}, and task decomposition \cite{pourreza-rafiei-2024-dts, pourreza2025reasoning} further enhance performance. Nevertheless, both fine-tuning and these extensions still often generate incomplete or invalid SQL queries. To address this limitation, we incorporate reinforcement learning with execution feedback, enabling self-correction and improving query validity.
% Complementing these approaches, we adopt a reinforcement learning approach with execution feedback, enabling self-correction and improve query validity.

\section{Methodology}
\subsection{Schema Retriever} \label{subsec:schema_linking}
To enable efficient schema linking with lightweight models, we adopt a retrieval-based approach in which candidate columns are retrieved from a vector database.

Each column is encoded into a dense embedding derived from its schema metadata (column name, description, table name, and value description). These embeddings are pre-computed and indexed in a vector database. At inference time, the question is embedded and compared against stored columns to retrieve the top-$k$ candidates by cosine similarity.

Effective retrieval requires not only general semantic similarity but also precise alignment between the question and the relevant columns. We therefore fine-tune the embedding model with a supervised contrastive objective. Unlike the standard supervised contrastive (SupCon) loss~\cite{khosla2020supervised}, which contrasts positives with all defined negatives based on semantic similarity, our setting further distinguishes columns that are semantically similar but irrelevant for SQL generation.

\begin{figure}[!t]
    \centerline{\includegraphics[width=1\linewidth]{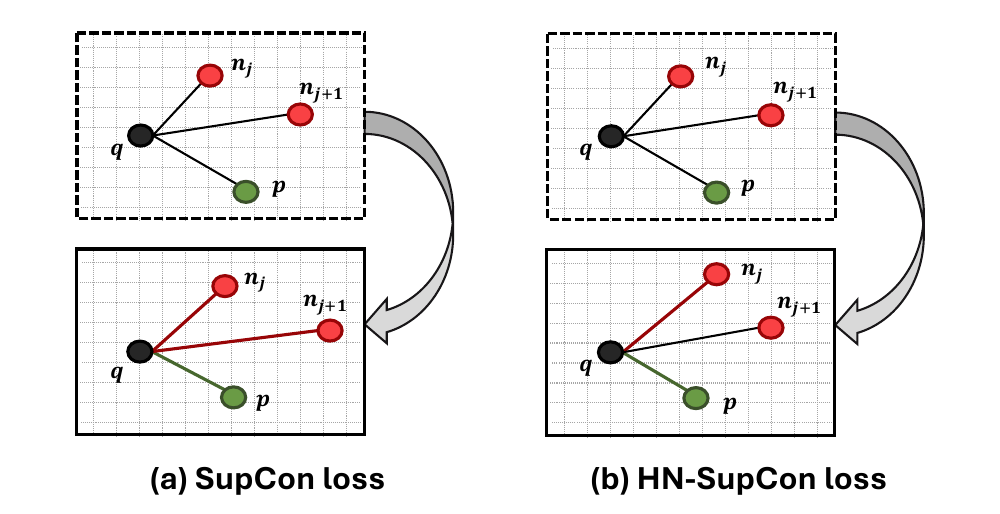}}
    \caption{Conceptual illustration of the difference between supervised contrastive (SupCon) loss and hard-negative filtered supervised contrastive (HN-SupCon) loss.}
    \label{fig:method-hnsupconloss}
\end{figure}

We observed that pre-trained models already assign low similarity to clearly irrelevant columns, making them easy negatives. The main challenge lies in hard negatives—columns that appear semantically related but are not required for SQL generation. To address this, we introduce a hard-negative filtered supervised contrastive loss (HN-SupCon), illustrated in Figure~\ref{fig:method-hnsupconloss}. Inspired by recent work that masks semantically related in-batch samples to avoid false negatives~\cite{zhang2025qwen3}, HN-SupCon selectively updates only hard negatives and ignores trivial ones.

Formally, given a question embedding $q_i$, its positive column $p_i$, and negatives $\{n_{ij}\}_{j}^{N_i}$, the loss is defined as:
\begin{equation}
    \begin{split}
    \mathcal{L}_{\mathrm{HN-SupCon}} 
    &= -\frac{1}{B} \sum_{i}^{B} \log \Bigg(
        \frac{ e^{s(q_i, p_i) / \tau} }
             { Z_i }
        \Bigg)
    \end{split}
\end{equation}

\begin{equation}
    Z_i = e^{s(q_i, p_i) / \tau}
         + \sum_{j}^{N_i} m_{ij}\, e^{s(q_i, n_{ij}) / \tau}
\end{equation}
where $s(\cdot,\cdot)$ denotes cosine similarity and $\tau$ is a temperature parameter. The mask $m_{ij}$ filters easy negatives:
\begin{equation}
    m_{ij} =
    \begin{cases}
    1 & \text{if } q_i \odot n_{ij} \geq q_i \odot p_i - 0.1, \\
    0 & \text{otherwise}
    \end{cases}
\end{equation}
The margin $0.1$ is chosen empirically (Table~\ref{appendix:hnsupcon_margin} in Appendix). %This formulation emphasizes hard negatives and reduces false positives during retrieval.
For efficiency, we limit the number of negatives sampled per question, as large schema (e.g., in BIRD) contains many irrelevant columns (See Appendix~\ref{sec:appendix-retriever-hyperparameters}).
% The effect of this hyperparameter, which control $N_i$, is analyzed in Section~\ref{sec:results-analysis-sr}.

\subsection{SQL Generator} \label{subsec:sql_generator}
\textbf{Supervised Fine-tuning (SFT).} To enhance the SQL generation ability of the pre-trained model ($\theta$), we first train the model to learn a conditional mapping from a natural language question ($Q$) and schema information ($S$) to the corresponding SQL query ($SQL$). The schema information includes column data types, primary key constraints, descriptive metadata, representative samples, and foreign key relations. The training objective is:
\begin{equation} \label{eq:sft_stage}
  \mathcal{L_{\mathrm{SFT}}}(\theta) = -\log P(SQL \mid Q, S; \theta)
\end{equation}

\begin{figure}[!t]
    \centering
    \includegraphics[width=0.9\linewidth]{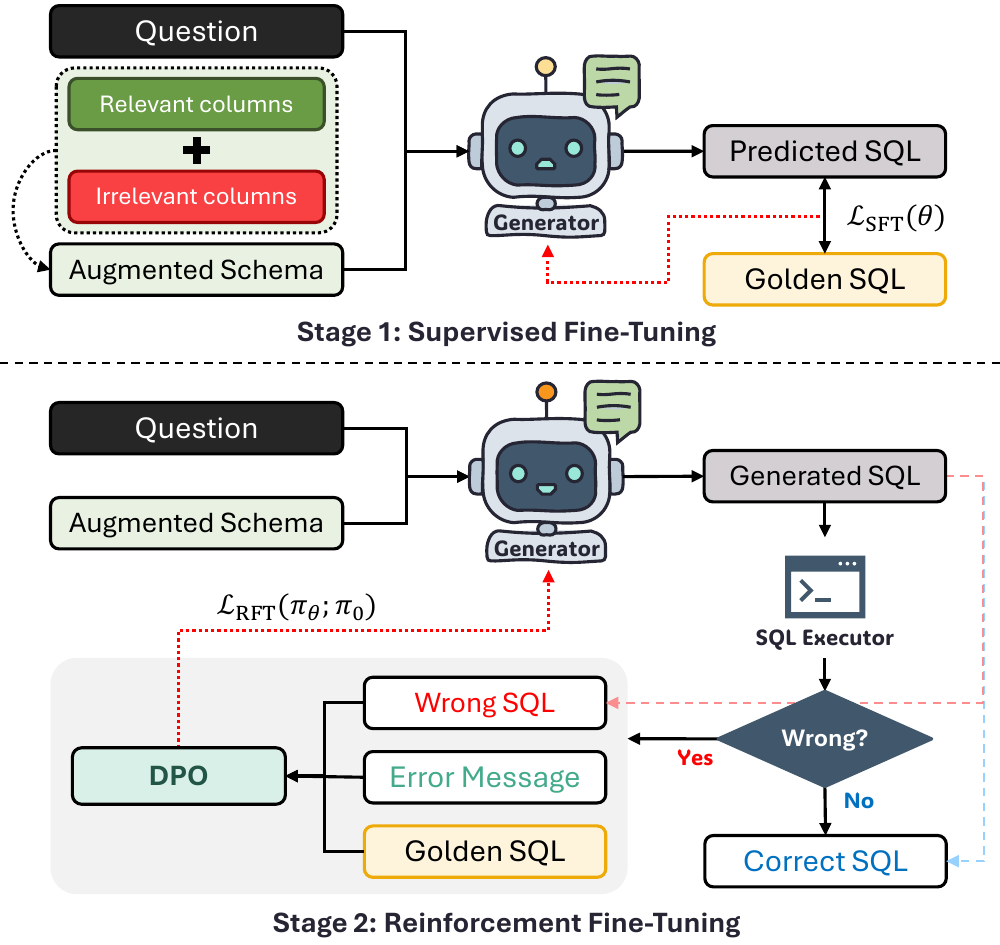}
    \caption{Two-stage training strategy of SQL Generator.}
    \label{method-generator}
\end{figure}

During inference, the model utilizes a fixed number $k$ of schema information retrieved by the Schema Retriever. However, training examples often contain fewer than $k$ schema information. To ensure consistency between training and inference, we augment training inputs by including only the schema elements explicitly referenced in the gold SQL, and filling the remaining slots with randomly sampled irrelevant schema information until reaching the fixed length $k$ (Figure~\ref{method-generator}). This approach allows the model to learn SQL generation in the presence of irrelevant but plausible schema information, which simulates potential false positives returned by the retriever at inference time while maintaining a consistent input structure. The value of $k$ is determined empirically based on dataset characteristics (Section~\ref{sec:results-analysis-sr}).

\textbf{Reinforcement Fine-tuning (RFT).} After the model reliably produces syntactically valid SQL, we train it to revise outputs using execution feedback. Specifically, we employ Direct Preference Optimization (DPO) \cite{rafailov2023direct} to align model behavior with preference signals derived from query execution results.

\begin{figure*}[th]
    \centering
    \includegraphics[width=0.9\linewidth]{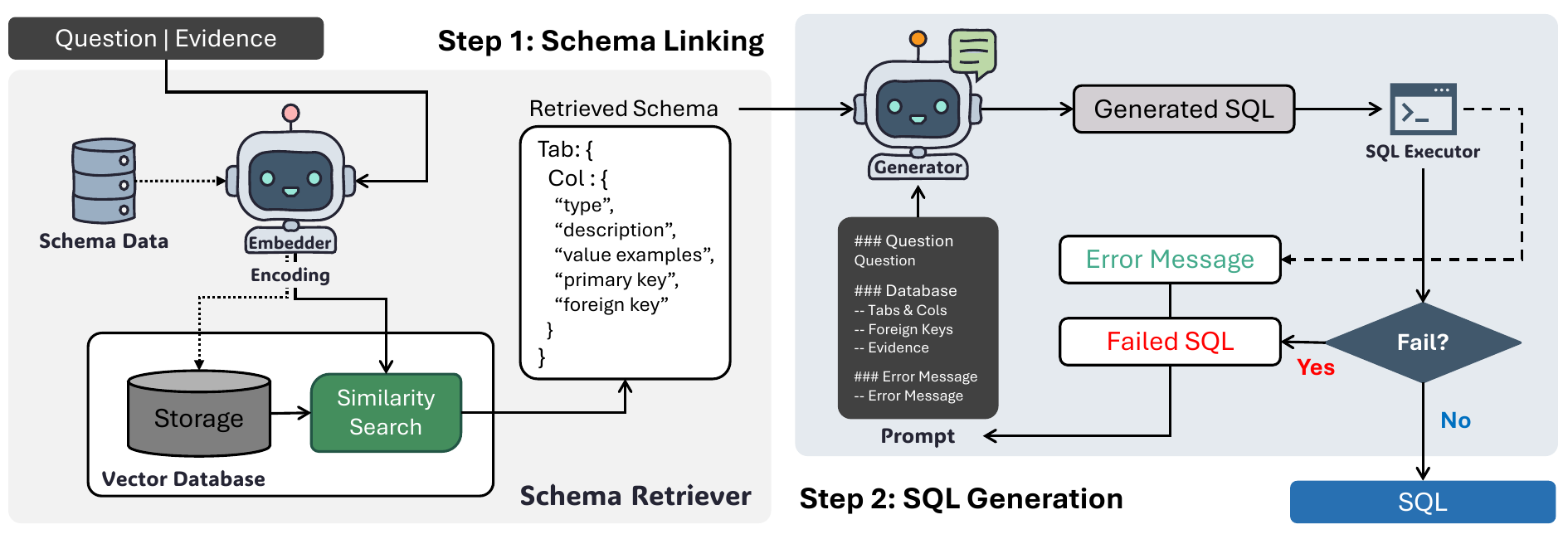}
    \caption{The inference pipeline of LitE-SQL.}
    \label{method-inference}
\end{figure*}

DPO operates on pairwise comparisons $\mathcal{D}=\left\{{x_i, y_i^w, y_i^l}\right\}_{i=1}^{N}$, where $x_i$ is the input, and $y_i^w$ and $y_i^l$ represent two generated queries with $y^w$ preferred over $y^l$. Instead of human-annotated preference, we generate multiple candidate queries per instance and label them by execution: successfully executing queries are preferred over failed ones. For failed queries, we also collect feedback (error messages) and incorporate it into the inputs used during SFT as illustrated in Figure~\ref{method-generator}.

To ensure data quality, when only one failed query exists, the corresponding ground truth query is treated as the preferred output. If multiple failed cases are present, we randomly sample non-overlapping successful outputs as preferred, ensuring the ground truth appears at least once. When successful queries are insufficient, we replicate the ground truth query to maintain balance.

Given the constructed dataset, the model is optimized using the following training objective:
\begin{equation}
    \begin{split}
    &\mathcal{L}_{\mathrm{RFT}}(\pi_\theta; \pi_{0}) \\
    &= \mathcal{L}_{\mathrm{DPO}}(y_i^w, y_i^l | x_i) + \alpha \mathcal{L}_{\mathrm{NLL}}(y_i^w | x_i) \\
    &= - \log \sigma \left( \beta \log \frac{\pi_\theta(y_i^w | x_i)}{\pi_{0}(y_i^w | x_i)} - \beta \log \frac{\pi_\theta(y_i^l | x_i)}{\pi_{0}(y_i^l | x_i)} \right) \\
    &\quad - \alpha \log \pi_\theta(y_i^w | x_i)
    \end{split}
\end{equation}
where $\pi_{0}$ is a fixed reference model initialized from $\pi_{\theta}$. The hyperparameter $\alpha$ controls the weight of the negative log-likelihood loss, and $\beta$ regulates divergence of $\pi_{\theta}$ from $\pi_{0}$.

\subsection{Inference}
We now describe how LitE-SQL operates at inference time (Figure~\ref{method-inference}). We assume preprocessing is completed: (i) database schemas are embedded and stored in a vector database using the fine-tuned embedding model (Embedder), and (ii) the SQL Generator has been trained through the two-stage process.

Given a user question (optionally with evidence)\footnote{Evidence annotations are unavailable in Spider 1.0.}, the input is encoded by Embedder and compared with stored schema embeddings via cosine similarity. The Schema Retriever then retrieves the top-$k$ candidate columns together with column types, descriptions, value examples, and key indicators (e.g., primary/foreign key status). The SQL Generator produces an initial SQL query conditioned on the question and retrieved schema, which is then executed against the target database. 

If execution fails, the system gathers feedback—including failed SQL query, corresponding error message, original question, and retrieved schema context—and feeds it back to the model for revision. During training, this execution feedback is incorporated into the model input and used to construct preference pairs for optimization. At inference time, the same feedback signal is used iteratively for execution-guided self-correction, with the detailed input format shown in Table~\ref{tab:input_template_rft}. The process repeats until a valid SQL query is produced or a predefined maximum number of iterations is reached, after which the final executable SQL is returned.

\section{Experimental Setup}
\subsection{Dataset}
We evaluate our model on two widely-used Text-to-SQL benchmarks: BIRD \cite{li2023can} and Spider 1.0 \cite{yu-etal-2018-spider}, both designed for cross-domain generalization. Specifically, Spider 1.0 includes 200 databases spanning 138 domains, while BIRD covers 95 large-scale databases from 37 professional domains. Both benchmarks feature complex SQL queries that involve operations such as \texttt{JOIN}, \texttt{GROUP BY}, and \texttt{HAVING}. Detailed dataset descriptions are provided in Appendix~\ref{sec:datasets}.
% Effective performance therefore requires models to comprehend database schemas, including table and column names as well as inter-table relationships. 

\subsection{Baselines}
We compare LitE-SQL against several recent methods on BIRD and Spider 1.0, covering both in-context learning and fine-tuning paradigms. Full baseline details are provided in Appendix~\ref{sec:baselines}.

\begin{table*}[!t]
    \centering
    \resizebox{1\linewidth}{!}{%
        \begin{tabular}{cllccc}
            \Xhline{1.0pt}
            \multicolumn{2}{c}{\multirow{2}{*}{\textbf{Method}}} & \multirow{2}{*}{\textbf{Generator}} & \multirow{2}{*}{\textbf{Size}} & \textbf{BIRD (Dev)} & \textbf{Spider 1.0 (Test)}\\ 
            & & & & \textbf{EX(↑)} & \textbf{EX (↑)} \\
            \Xhline{0.5pt}
            \multirow{11}{*}{In-context Learning}
            & ChatGPT + CoT \cite{li2023can}                       & GPT-3.5        & 175B  & 36.64          & -\\ 
            & DIN-SQL \cite{pourreza2023din}                       & GPT-4          & 175B  & 50.72          & 85.30\\
            & DAIL-SQL \cite{10.14778/3641204.3641221}             & GPT-4          & 175B  & 54.76          & 86.60\\
            & TA-SQL \cite{qu-etal-2024-generation}                & GPT-4          & 175B  & 56.19          & -\\
            & SuperSQL \cite{10.14778/3681954.3682003}             & GPT-4          & 175B  & 58.50          & -\\
            & MAC-SQL \cite{wang-etal-2025-mac}                    & GPT-4          & 175B  & 59.39          & 86.75\\
            & MCS-SQL \cite{lee-etal-2025-mcs}                     & GPT-4          & 175B  & 63.36          & \textbf{89.60}\\
            & CHESS \cite{DBLP:journals/corr/abs-2405-16755}       & Gemini 1.5 Pro & 200B  & 68.31          & 87.20\\
            & Distillery \cite{maamari2024the}                     & GPT-4o         & 200B  & 67.21          & -\\
            & CHASE-SQL \cite{pourreza2025chasesql}                & Gemini 1.5     & 200B  & \textbf{73.01} & 87.60\\
            \Xhline{0.5pt}
            \multirow{5}{*}{Fine-tuning} & \multirow{1}{*}{CodeS \cite{li2024codes}} & \multirow{1}{*}{StarCoder} 
               & 7B  & 57.17 & -\\
            & \multirow{1}{*}{OmniSQL \cite{li2025omnisql}} & \multirow{1}{*}{Qwen2.5-Coder} 
                & 7B   & 63.90 & 87.90\\
            & DTS-SQL \cite{pourreza-rafiei-2024-dts} & DeepSeek & 7B & 55.80 & 77.10\\ 
            & \multirow{1}{*}{Reasoning-SQL \cite{pourreza2025reasoning}} & \multirow{1}{*}{Qwen2.5-Coder} 
                & 7B   & 68.05 & 78.72\\
            & \multirow{1}{*}{\textbf{LitE-SQL}} & \multirow{1}{*}{Qwen2.5-Coder} 
                & 7B    & \underline{72.10} & \underline{88.45} \\ 
            \Xhline{1.0pt}
            \multicolumn{4}{l}{\footnotesize{The performance of the baselines is obtained from the corresponding paper.}}
        \end{tabular}
    }
    \caption{EX (\%) of our model with correct schema and comparison with baselines on BIRD development set and Spider 1.0 test set. Among fine-tuning–based baselines with 7B parameters, LitE-SQL achieves the best performance. \textbf{Bold} numbers indicate the best performance, and \underline{underlined} numbers indicate the second-best.}
    \label{tab:overall-performance-part}
\end{table*}

\subsection{Metrics}
% We evaluate model performance using the following metrics.
\textbf{Execution Accuracy (EX).} Following the standard in BIRD and Spider 1.0, we adopt EX as the primary evaluation metric. EX measures whether the predicted SQL query yields the same execution result as the ground-truth query on the target database.
\begin{equation}
    \mathrm{EX} = \frac{1}{N} \sum_{n=1}^{N} \mathbb{I}(V_n = \hat{V}_n),
    \label{eq:ex_def}
\end{equation}
where $V_n$ denotes the result set returned by executing the ground-truth query, and $\hat{V}_n$ is the result set from the predicted query. The indicator function $\mathbb{I}(\cdot)$ is defined as:
\begin{equation}
    \mathbb{I}(V, \hat{V}) = 
    \begin{cases}
    1, & \text{if } V = \hat{V}, \\
    0, & \text{otherwise}.
    \end{cases}
    \label{eq:indicator}
\end{equation}

\textbf{True Positive Rate (TPR).} The proportion of ground-truth relevant columns that are retrieved by the Schema Retriever. Higher values indicate more effective retrieval of necessary columns.
\begin{equation}
    \text{TPR} = \frac{\text{\# of retrieved relevant columns}}{\text{\# of ground-truth relevant columns}}
\end{equation}

\textbf{False Positive Rate (FPR).} The proportion of retrieved columns that are not part of the ground-truth relevant columns. Lower values indicate better precision in schema retrieval.
\begin{equation}
    \text{FPR} = \frac{\text{\# of retrieved irrelevant columns}}{\text{\# of retrieved columns}}
\end{equation}

\textbf{Schema Linking Recall (SLR).} The proportion of queries where all relevant columns are successfully retrieved. We consider a case successful if the Schema Retriever retrieves the complete set of ground-truth relevant columns. Higher SLR indicates stronger schema linking performance.
\begin{equation}
    \text{SLR} = \frac{\text{\# of successful cases}}{\text{\# of total queries}}
\end{equation}

\subsection{Implementation Details}
We employ ChromaDB\footnote{https://github.com/chroma-core/chroma} as the vector database for lightweight integration, in-memory retrieval, and cosine similarity search. The embedding backbone for Schema Retriever is Qwen3-0.6B-Embedding~\cite{zhang2025qwen3}, fine-tuned with the HN-SupCon loss (Section~\ref{subsec:schema_linking}). We adopt Qwen2.5-Coder models \cite{hui2024qwen2} with 1.5B, 3B, and 7B parameters for SQL Generator. All models are fine-tuned on 4 A100 GPUs with the Huggingface library \cite{wolf2019huggingface}, initialized from publicly released checkpoints\footnote{https://huggingface.co/Qwen}. Detailed hyperparameter settings are provided in Appendix~\ref{sec:hyperparameter}.
% \textbf{Schema Retriever.} We employ ChromaDB\footnote{https://github.com/chroma-core/chroma} as the vector database for lightweight integration, in-memory retrieval, and cosine similarity support. The embedding backbone is Qwen3-0.6B-Embedding~\cite{zhang2025qwen3}, fine-tuned with the HN-SupCon loss (Section~\ref{subsec:schema_linking}).

% \textbf{SQL Generator.} We adopt Qwen2.5-Coder models \cite{hui2024qwen2} with 1.5B, 3B, and 7B parameters. All models are fine-tuned on 4 A100 GPUs with the Huggingface library \cite{wolf2019huggingface}, initialized from publicly released checkpoints\footnote{https://huggingface.co/Qwen}. 

% Detailed hyperparameter settings are provided in Appendix~\ref{sec:hyperparameter}.

\section{Results and Analysis}
\subsection{Baselines Comparison}
We evaluate LitE-SQL on the BIRD development set\footnote{Since the test set is not publicly available, we follow prior work and report performance on the development set.} and the Spider 1.0 test set. Recent Text-to-SQL systems typically perform schema linking in advance, using LLMs to extract relevant schema information. To ensure a controlled comparison, we adopt the same setting and provide the resulting schema information during SQL generation.

We first compare LitE-SQL against in-context learning methods. As shown in Table~\ref{tab:overall-performance-part}, most methods rely on 175B-200B \cite{fello2024} parameter models, whereas our fine-tuned 7B model achieves 72.10\% on BIRD and 88.45\% on Spider 1.0, outperforming the majority of GPT-4 and Gemini-1.5 based methods. Although CHASE-SQL surpasses our method by 0.91\% on BIRD, we outperform it by 0.85\% on Spider 1.0. Compared to MCS-SQL, which achieves the highest Spider 1.0 score, our method yields a 1.15\% gap while exceeding it by 8.74\% on BIRD. These results underscore the generalization capability of LitE-SQL, offering competitive performance with significantly fewer parameters.

Within the fine-tuning category, LitE-SQL consistently outperforms prior 7B-scale baselines, with average improvements of 10.87\% on BIRD and 7.21\% on Spider 1.0. The consistent improvements across benchmarks demonstrate that the two-stage training strategy efficiently utilizes model capacity, balancing predictive performance and computational efficiency.

\subsection{Ablation Study}
\begin{table}[!t]
\renewcommand{\arraystretch}{1}
    \centering
    \resizebox{\linewidth}{!}{
        \begin{tabular}{cccll}
        \Xhline{1.0pt}
        {\multirow{2}{*}{\textbf{Retriever}}} & \multicolumn{2}{c}{\textbf{Generator}} & \multicolumn{2}{c}{\textbf{EX (↑)}} \\ 
        & \textbf{SFT} & \textbf{RFT} & \textbf{BIRD (Dev)} & \textbf{Spider 1.0 (Test)}\\
        \Xhline{0.5pt}
        \textcolor{red}{\ding{55}}       & \textcolor{red}{\ding{55}}       & \textcolor{red}{\ding{55}}       & 39.31                   & 61.61 \\
        \textcolor{darkgreen}{\ding{51}} & \textcolor{red}{\ding{55}}       & \textcolor{red}{\ding{55}}       & 43.16 (+3.85)           & 64.28 (+2.67) \\
        \textcolor{darkgreen}{\ding{51}} & \textcolor{darkgreen}{\ding{51}} & \textcolor{red}{\ding{55}}       & 58.21 (+18.90)          & 83.56 (+21.95) \\
        \textcolor{darkgreen}{\ding{51}} & \textcolor{darkgreen}{\ding{51}} & \textcolor{darkgreen}{\ding{51}} & \textbf{60.56 (+21.25)} & \textbf{84.35 (+22.74)}\\
        \Xhline{1.0pt}
        \end{tabular}
    }
    \caption{Overall result of ablation study.}
    \label{tab:overall_ablation_study}
\end{table}

\begin{table}[!t]
\renewcommand{\arraystretch}{1}
    \centering
    \resizebox{\linewidth}{!}{
    \begin{tabular}{lccccc}
        \Xhline{1.0pt}
        \textbf{Method} & \textbf{Size} & \textbf{TPR (↑)} & \textbf{FPR (↓)} & \textbf{SLR (↑)} & \textbf{EX (↑)} \\ \Xhline{0.5pt}
        \begin{tabular}[c]{@{}l@{}}CodeS\\ \cite{li2024codes}\end{tabular}                       & 3.5B & 89.64          & 74.16         & 65.31          & 51.70 \\  
        \begin{tabular}[c]{@{}l@{}}CHESS\\ \cite{DBLP:journals/corr/abs-2405-16755}\end{tabular} & 200B & 87.15          & \textbf{8.27} & 61.9           & \textbf{57.14} \\ 
        \textbf{LitE-SQL}                                                                                 & 0.6B & \textbf{95.23} & 80.28         & \textbf{82.31} & 56.46 \\ 
        \Xhline{1.0pt}
    \end{tabular}
    }
    \caption{Performance of schema linking on subsampled-BIRD compared to baselines.}
    \label{tab:performance_sl_compared_baselines}
\end{table}

To evaluate the contributions of each component, we conduct a comprehensive ablation study, with results presented in Table~\ref{tab:overall_ablation_study}~\footnote{Unless otherwise specified, all reported results in subsequent sections are based on the 7B model.}. 

We first assess the effect of the Schema Retriever. In the baseline configuration, the retriever is disabled, and the complete database schema is fed directly to the SQL Generator without fine-tuning. This leads to a notable performance degradation compared to the retriever-enabled setup, highlighting the importance of filtering irrelevant schema elements to constrain the generator's search space, thereby underscoring the necessity of schema linking.

We then evaluate the two-stage training strategy. After the SFT stage, the model outperforms the prompt-engineering baseline by more effectively leveraging retrieved schema information for SQL generation. Importantly, this improvement should not be interpreted as diminishing the role of retrieval. Instead, SFT serves as a prerequisite that enables the model to fully benefit from schema retrieval. As shown in Table~\ref{tab:performance_sl_compared_baselines}, once the generator has been sufficiently trained, retrieval quality becomes a decisive factor. For instance, CodeS with weaker schema linking imposes a clear performance ceiling, limiting the effectiveness of the generator.

Additional gains are achieved from the RFT stage, where execution feedback is used to revise SQL outputs. Although improvements over SFT are modest, RFT enhances robustness by reducing both semantic and syntactic errors. Detailed error analysis is provided in §\ref{subsec: analysis_generator}.

Overall, the ablation results reveal a clear stage-dependent interaction among schema retrieval, SFT, and RFT. SFT delivers the primary performance gains, schema retrieval constrains the generator’s search space once training is in place, and RFT further improves robustness. Their combination produces cumulative improvements on both BIRD and Spider 1.0, validating the complementary roles of all components in the proposed framework.

\subsection{Analysis of Schema Retriever} \label{sec:results-analysis-sr}
\begin{table}[!t]
    \centering
    \resizebox{\linewidth}{!}{
        \begin{tabular}{lccc}
        \Xhline{1.0pt}
        Model & \textbf{TPR (↑)} & \textbf{FPR (↓)} & \textbf{SLR (↑)} \\ 
        \Xhline{0.5pt}
        Qwen3-Embedding-0.6B& 82.80          & 83.62          & 59.91 \\
        + SupCon            & 95.03          & 81.04          & 82.46 \\
        + \textbf{HN-SupCon}         & \textbf{96.71} & \textbf{80.71} & \textbf{87.42} \\ 
        \Xhline{1.0pt}
        \end{tabular}
    }
    \caption{Effect of loss functions on schema retrieval performance on BIRD development set.}
    \label{tab:analysis-hn-supconloss-bird}
\end{table}

\begin{table}[!t]
    \centering
    \resizebox{\linewidth}{!}{
        \begin{tabular}{lccc}
        \Xhline{1.0pt}
        Model & \textbf{TPR (↑)} & \textbf{FPR (↓)} & \textbf{SLR (↑)} \\ 
        \Xhline{0.5pt}
        Qwen3-Embedding-0.6B& 99.22          & 84.08          & 96.88 \\
        + SupCon            & 96.32          & 84.12          & 96.13 \\
        + \textbf{HN-SupCon}         & \textbf{99.27} & \textbf{83.96} & \textbf{99.16} \\
        \Xhline{1.0pt}
        \end{tabular}
    }
    \caption{Effect of loss functions on schema retrieval performance on Spider 1.0 test set.}
    \label{tab:analysis-hn-supconloss-spider}
\end{table}
\textbf{Schema linking baselines.} In this experiment, we evaluate schema linking performance against representative baselines. For fairness, we use each method’s schema linking strategy to extract relevant columns, while employing our SQL Generator for execution. Following prior work~\cite{DBLP:journals/corr/abs-2405-16755, maamari2024the}, evaluation is conducted on a subsampled BIRD development set (10\% per database, 147 examples). We compare against CodeS and CHESS, two widely adopted baselines~\cite{maamari2024the, pourreza2025chasesql, pourreza2025reasoning, 10.14778/3681954.3682003}.

As shown in Table~\ref{tab:performance_sl_compared_baselines}, SLR plays a crucial role in determining EX. Despite a higher FPR than CodeS, LitE-SQL’s stronger SLR yields a 4.76\% gap in EX, showing that higher recall can compensate for moderate false positives. By contrast, when the FPR gap is large, as between CHESS and LitE-SQL, excessive irrelevant columns reduce EX. The EX difference between CHESS and LitE-SQL is modest, only a 0.68\% gap, yet notable given their difference in model scale and architecture: CHESS relies on a 200B-parameter multi-agent LLM with chain-of-thought reasoning, while LitE-SQL achieves comparable accuracy with a 0.6B-parameter retriever.

\textbf{Effect of loss functions.} We evaluate the impact of different training objectives on schema retrieval by comparing three settings: the vanilla embedding model (Qwen3), fine-tuned with SupCon loss (Qwen3 + SupCon), and fine-tuned with our proposed HN-SupCon loss (Qwen3 + HN-SupCon). As shown in Table~\ref{tab:analysis-hn-supconloss-bird} and Table~\ref{tab:analysis-hn-supconloss-spider}, HN-SupCon achieves the best trade-off, yielding higher TPR and SLR while reducing FPR. These results confirm that emphasizing hard negatives enables the model to distinguish columns that are semantically similar but functionally irrelevant, thereby mitigating false positives and improving retrieval quality.

\begin{figure}[!t]
    \includegraphics[width=1\linewidth]{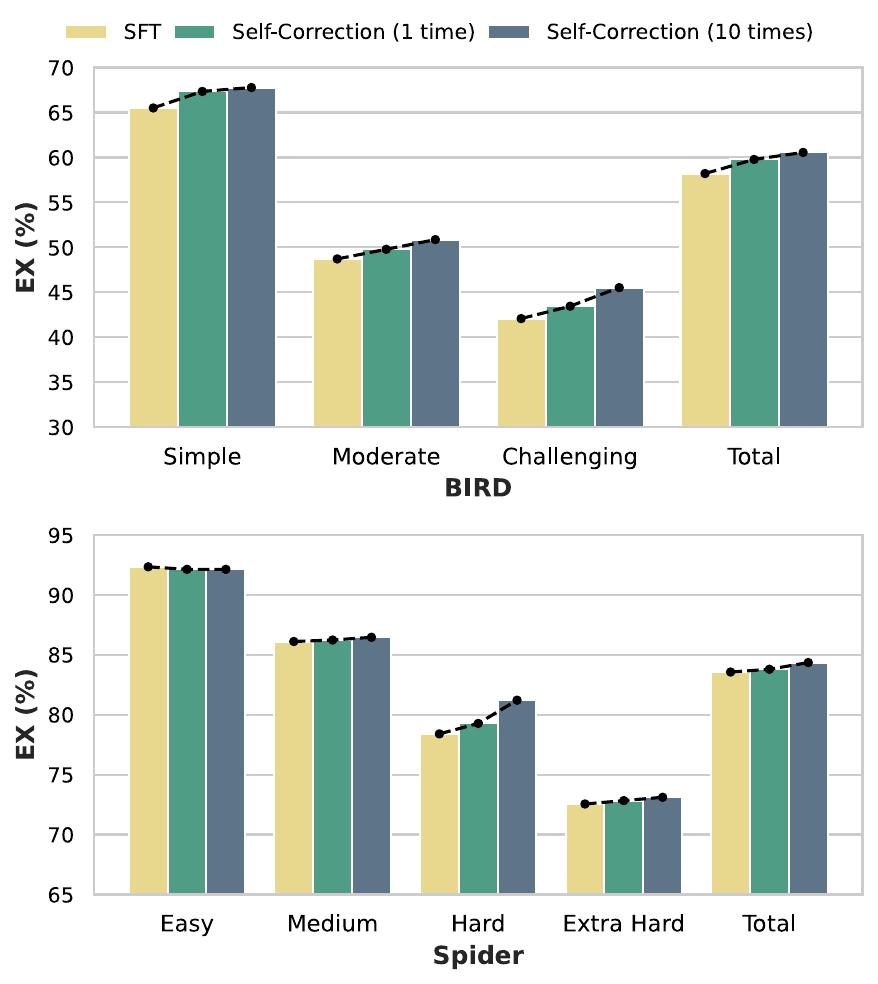}
    \caption{Effect of the number of self-correction iterations across different question difficulty.}
    \label{analysis_genetor}
\end{figure}
\textbf{Effect of hyperparameters.} We analyze the impact of two key hyperparameters in fine-tuning the embedding and filtering models: the number of negative samples and the number of retrieved columns, denoted as $N_i$ and $k$, respectively. Because large schemas contain many irrelevant columns, we limit the number of negatives sampled per question. On both datasets, setting $N_i=8$ yields the best trade-off between FPR and SLR, while using $k=25$ retrieved columns achieves the highest SLR. Detailed results of corresponding analysis on datasets are provided in Appendix~\ref{sec:appendix-retriever-hyperparameters}.

\subsection{Analysis of SQL Generator} \label{subsec: analysis_generator}
\begin{figure}[!t]
    \includegraphics[width=1\linewidth]{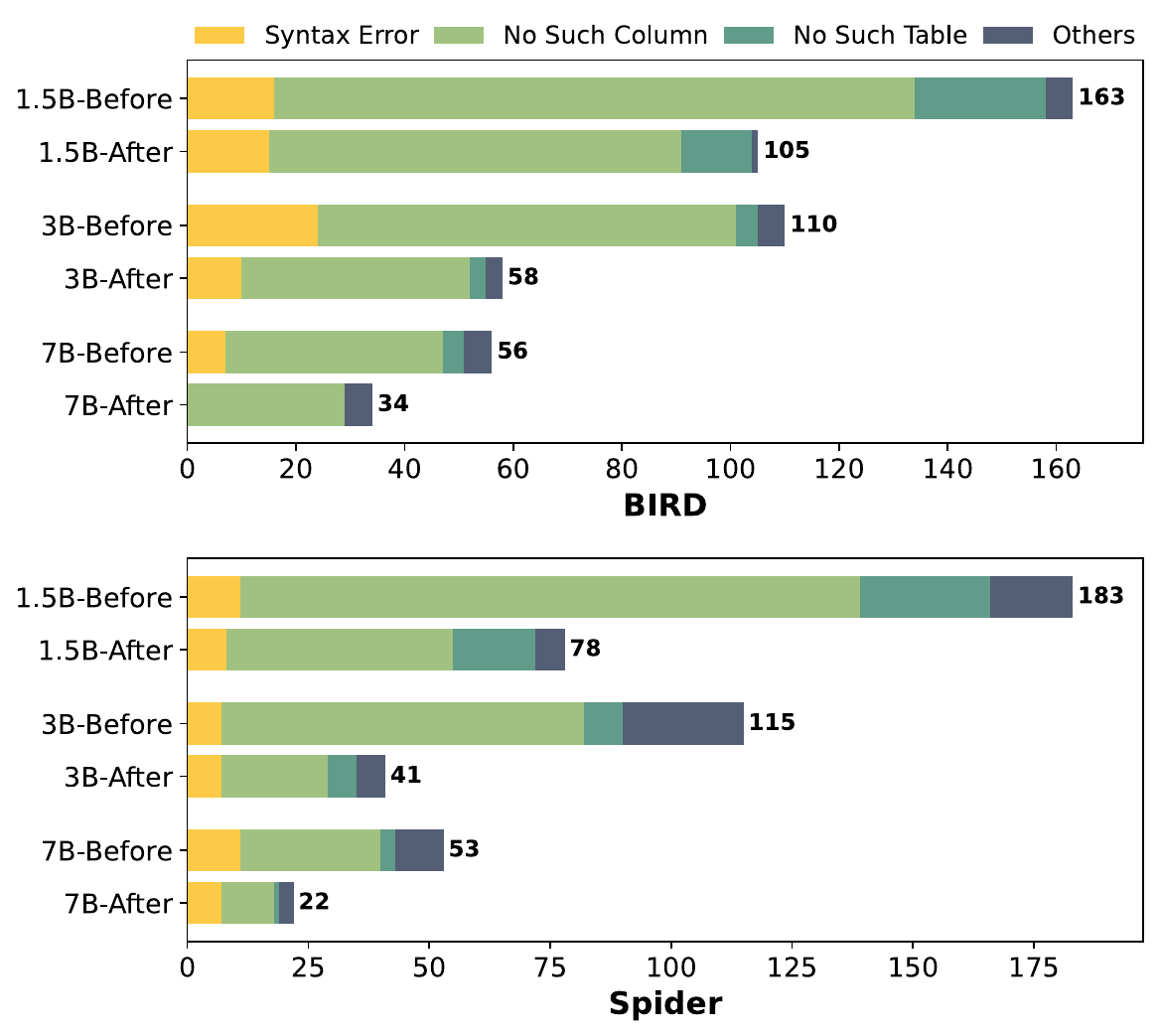}
    \caption{The distribution of execution error types before and after self-correction for each model size. The x-axis indicates the number of errors, and the number of each bar denotes the total errors.}
    \label{analysis_correction}
\end{figure}
\textbf{Effect of self-correction iterations.} We evaluate the effectiveness of self-correction by analyzing both the number of correction iterations and their impact across different question difficulty levels, as illustrated in Figure~\ref{analysis_genetor}. Our results demonstrate that incorporating executor feedback during generation consistently enhances EX across all correction rounds. This confirms the utility of the self-correction ability acquired during the RFT stage. Furthermore, the first self-correction pass yields the largest accuracy gain on the BIRD benchmark. While subsequent passes continue to improve performance, the gains diminish progressively, suggesting that most correctable errors are addressed early, with later iterations refining a small subset of harder cases.

\textbf{Effect of execution feedback.}
To assess how the generator leverages executor feedback during self-correction, we categorize execution errors into four types: \textit{syntax errors}, \textit{no such column}, \textit{no such table}, and \textit{other} (miscellaneous errors). We compare the distribution of these error categories before and after self-correction to quantify improvements. As shown in Figure~\ref{analysis_correction}, self-correction consistently reduces the frequency of all error categories across model scales.

Semantic errors—particularly \textit{no such column} and \textit{no such table}—are of notable importance, as they reflect misalignment between the natural language question and the underlying database schema. Among these, \textit{no such column} error, initially the most prevalent, decreases substantially after correction. Moreover, \textit{no such table} errors are almost entirely eliminated in larger models. These results suggest that self-correction significantly improves semantic alignment, with larger models benefiting the most. \textit{Syntax errors} are observed across all model sizes prior to correction but are reduced afterward, with the 7B model achieving full resolution on the BIRD dataset.

Overall, these findings underscore the effectiveness of the RFT stage in enhancing model robustness by mitigating both semantic and syntactic errors.

\section{Conclusion}
In this paper, we present LitE-SQL, a lightweight and efficient framework for Text-to-SQL with vector-based schema linking and execution-guided self-correction. We evaluated LitE-SQL on two representative datasets, showing that it matches or outperforms LLM-based baselines in EX while requiring up to \mbox{$30\times$} fewer parameters. Beyond empirical gains, LitE-SQL can serve as a privacy-preserving and resource-efficient solution suitable for on-premises deployment.
% We presented LitE-SQL, a lightweight and efficient framework for Text-to-SQL generation. The framework comprises two components: a Schema Retriever, which retrieves relevant schema information via vector-based similarity search, and a SQL Generator, trained through a two-stage process of supervised fine-tuning followed by execution-guided reinforcement learning.

% Comprehensive evaluations on the BIRD and Spider 1.0 demonstrate that LitE-SQL achieves execution accuracy on par with or surpassing LLM-based baselines, while requiring up to \mbox{$30\times$} fewer parameters. Ablation studies further validate the effectiveness of efficient schema retrieval and execution-guided self-correction. 

% Beyond empirical gains, LitE-SQL offers a privacy-preserving and resource-efficient solution suitable for on-premise deployment, effectively addressing key limitations of proprietary LLMs.

\section*{Limitations}
While LitE-SQL achieves competitive performance, there remain two areas for improvement. 

First, retrieving a fixed $k$ columns inherently increases the likelihood of including true positives, but at the same time inevitably introduces false positives. While our design partially mitigates this trade-off through embedding space optimization, reducing residual false positives remains a promising avenue for future improvement.
% First, in the schema linking step, the ratio of false positives (i.e., irrelevant columns retrieved as relevant) is still non-negligible. Although EX is comparable to recent work, reducing false positives could yield further gains in overall performance. 

Second, while the current self-correction mechanism improves robustness by handling execution failures, it is mainly effective for syntactic errors that can be detected during execution. This design already mitigates a large portion of query failures, but semantically inaccurate yet executable queries—for instance, those joining the wrong table but still returning results—remain unresolved. Developing mechanisms that capture such semantic errors represents a natural extension of our framework.
% Second, while the current self-correction mechanism improves robustness by handling execution failures, it primarily focuses on syntactic errors. An interesting extension would be to also capture and correct semantically inaccurate but executable queries, which could further enhance overall performance. We leave such directions for future work.
% While LitE-SQL achieves competitive performance, it has two main limitations: (i) reducing the ratio of irrelevant columns in retrieved results by addressing the top-ranked selection limitation of the Schema Retriever, and (ii) enhancing the SQL Generator’s self-correction capability to identify and correct semantically inaccurate queries that are executable but produce erroneous results.

\section*{Acknowledgments}
This research was supported by the National Research Foundation (NRF) and IITP grants funded by the Korean Government (MSIT) (No. RS-2023-00229822 and No. RS-2020-II201361, Artificial Intelligence Graduate School Program at Yonsei University).

\bibliography{latex/references}

\appendix
\section{Detailed Experimental Setup}
\subsection{Datasets} \label{sec:datasets}
\textbf{BIRD}\footnote{https://bird-bench.github.io/} The BIRD benchmark aims to bridge the gap between academic research and real-world applications by encompassing 37 professional domains that present practical challenges, including noisy or incomplete database values, external knowledge integration, and SQL efficiency on large databases. It comprises questions categorized by difficulty: simple, moderate, and challenging, each paired with a SQL query, and annotated external knowledge for selected samples. \\
\textbf{Spider 1.0}\footnote{https://yale-lily.github.io/spider} Spider 1.0 is a large-scale, cross-domain Text-to-SQL benchmark designed to evaluate the generalization ability of models across diverse databases and query structures. Each question is paired with an SQL query and categorized into four difficulty levels: easy, medium, hard, and extra hard, depending on query complexity. The benchmark emphasizes model performance on unseen databases and SQL queries, aiming to assess generalization beyond pattern memorization.

Table~\ref{tab:statistic_database} summarizes key statistics of both datasets.

\subsection{Baselines} \label{sec:baselines}
\textbf{In-context Learning.} Prompt engineering is central to this paradigm. DAIL-SQL \cite{10.14778/3641204.3641221} selects examples by similarity to enhance relevance, while SuperSQL \cite{10.14778/3681954.3682003} introduces intermediate representations to improve semantic alignment. Distillery \cite{maamari2024the} demonstrates effective augmentation and correction when schema fit within the context window.

To handle complex questions, several models employ question decomposition. DIN-SQL \cite{pourreza2023din}, MAC-SQL \cite{wang-etal-2025-mac}, and CHASE-SQL \cite{pourreza2025chasesql} decompose questions into sub-questions, while MAC-SQL, ChatGPT + CoT \cite{li2023can}, and CHASE-SQL further integrate chain-of-thought prompting to enhance interpretability and accuracy. Pipeline-based approaches such as TA-SQL \cite{qu-etal-2024-generation} and MCS-SQL \cite{lee-etal-2025-mcs} explicitly perform schema linking to reduce errors and improve schema recall. Multi-agent framework, like MAC-SQL, DIN-SQL, CHESS \cite{DBLP:journals/corr/abs-2405-16755}, CHASE-SQL, and MCS-SQL, which modularize tasks such as schema linking, decomposition, and SQL refinement. \\
\textbf{Fine-tuning.} Fine-tuning language models on SQL corpora improves alignment between natural language and SQL. From a data perspective, CodeS \cite{li2024codes} incrementally pre-trains on curated SQL datasets with schema-aware retrieval,  while OmniSQL \cite{li2025omnisql} synthesizes SynSQL-2.5M, a large and diverse corpus designed for realistic query patterns.

Regarding training methodologies, DTS-SQL \cite{pourreza-rafiei-2024-dts} implements a two-stage fine-tuning process that first handles schema linking before SQL generation. Reasoning-SQL \cite{pourreza2025reasoning} integrates partial rewards such as schema linking accuracy and execution correctness via Group Relative Policy Optimization to foster intermediate reasoning.

\begin{table}[t]
\renewcommand{\arraystretch}{1}
    \centering
    \resizebox{\linewidth}{!}{%
        \begin{tabular}{ccccc}
            \Xhline{1.0pt}
            \textbf{Dataset} & \textbf{\# DB} & \textbf{\# Table/DB}& \textbf{\# Row/DB} \\ \Xhline{0.5pt}
            BIRD       & 95  & 7.3 & 549K \\
            Spider 1.0 & 200 & 5.1 & 2K \\
            \Xhline{0.5pt}
            \textbf{Dataset} & \textbf{\# Train} & \textbf{\# Valid} & \textbf{\# Test} \\
            \Xhline{0.5pt}
            BIRD       & 9376 & 1534 & -    & \\
            Spider 1.0 & 8659 & 1034 & 2147 & \\
            \Xhline{1.0pt}
        \end{tabular}
    }
    \caption{Database Statistics of BIRD and Spider 1.0.}
    \label{tab:statistic_database}
\end{table}

\subsection{Hyperparameters} \label{sec:hyperparameter}

\begin{table*}[!t]
    \centering
    \resizebox{1\linewidth}{!}{%
        \begin{tabular}{llccc}
            \Xhline{1.0pt}
            {\multirow{2}{*}{\textbf{Method}}} & \multirow{2}{*}{\textbf{Generator}} & \multirow{2}{*}{\textbf{Size}} & \textbf{BIRD (Dev)} & \textbf{Spider 1.0 (Test)}\\ 
            & & & \textbf{EX(↑)} & \textbf{EX (↑)} \\
            \Xhline{0.5pt}
            \multirow{4}{*}{CodeS \cite{li2024codes}} & \multirow{4}{*}{StarCoder} 
                & 1B  & 50.46 & -\\
            & & 3B  & 55.02 & -\\
            & & 7B  & 57.17 & -\\
            & & 15B & 58.47 & -\\ 
            \cdashline{1-5}[1pt/1pt] 
            \multirow{3}{*}{OmniSQL \cite{li2025omnisql}} & \multirow{3}{*}{Qwen2.5-Coder} 
                & 7B   & 63.90 & 87.90\\
            & & 14B  & 64.20 & \underline{88.30}\\
            & & 32B  & 64.50 & 87.60\\ 
            \cdashline{1-5}[1pt/1pt] 
            DTS-SQL \cite{pourreza-rafiei-2024-dts} & DeepSeek & 7B & 55.80 & 77.10\\ 
            \cdashline{1-5}[1pt/1pt]
            \multirow{2}{*}{Reasoning-SQL \cite{pourreza2025reasoning}} & \multirow{2}{*}{Qwen2.5-Coder} 
                & 7B   & 68.05 & 78.72\\
            & & 14B  & \textbf{72.29} & 81.43\\ 
            \cdashline{1-5}[1pt/1pt]  
            \multirow{3}{*}{\textbf{LitE-SQL}} & \multirow{3}{*}{Qwen2.5-Coder} 
                & 1.5B  & 65.19 & 83.00 \\
            & & 3B    & 70.01 & 85.28 \\
            & & 7B    & \underline{72.10} & \textbf{88.45} \\ 
            \Xhline{1.0pt}
            \multicolumn{4}{l}{\footnotesize{The performance of the baselines is obtained from the corresponding paper.}}
        \end{tabular}
    }
    \caption{Extended results for the fine-tuning baselines in Table~\ref{tab:overall-performance-part}.}
    \label{tab:overall-performance-fine}
\end{table*}

\begin{table}[t]
    \centering
    \resizebox{1\linewidth}{!}{%
        \begin{tabular}{ccccccc}
            \Xhline{1.0pt}
            \multirow{2}{*}{\textbf{\begin{tabular}[c]{@{}c@{}}Margin\\ Values\end{tabular}}} & \multicolumn{3}{c}{\textbf{BIRD (Dev)}} & \multicolumn{3}{c}{\textbf{Spider 1.0 (Test)}} \\ \cline{2-7} & \textbf{TPR (↑)} & \textbf{FPR (↓)} & \textbf{SLR (↑)} & \textbf{TPR (↑)} & \textbf{FPR (↓)} & \textbf{SLR (↑)} \\ 
            \Xhline{0.5pt}
            0   & 96.08          & 80.83          & 85.46          & 99.18          & 83.98          & 98.88 \\
            0.1 & \textbf{96.71} & \textbf{80.71} & \textbf{87.42} & \textbf{99.27} & \textbf{83.96} & \textbf{99.16} \\
            0.2 & 96.22          & 80.81          & 86.18          & 99.03          & 83.98          & 98.74 \\ 
            \Xhline{1.0pt}
        \end{tabular} 
    }
    \caption{Evaluation of margin values for HN-SupCon loss.}
    \label{appendix:hnsupcon_margin}
\end{table}

The hyperparameter configurations for the implementation are summarized as follows. \\
\textbf{Schema Retriever.} The embedding model is trained for 1 epoch using the AdamW optimizer with a learning rate of $5 \times 10^{-5}$ and temperature 0.07. Batch sizes are set to 16 for BIRD and 24 for Spider.\\
\textbf{SQL Generator (SFT).} Each model is trained for 2000 steps with a batch size of 16 using the AdamW optimizer. Learning rates are set to $1 \times 10^{-4}$ for the 1.5B and 3B models, and $5 \times 10^{-5}$ for the 7B model. A cosine learning rate scheduler with a warm-up ratio of $0.05$ is employed. We apply LoRA adaptation \cite{hu2022lora} with rank $r = 64$ for the 1.5B models, and $r = 128$ for the 3B and 7B models. The scaling factor is set as $\alpha_{\text{LoRA}} = 2r$, and dropout is fixed at $0.05$. \\
\textbf{SQL Generator (RFT).} Each model is further fine-tuned for 2 epochs using pairwise preference samples generated with a temperature of 0.7. The batch size remains 16. Learning rates are set to $5 \times 10^{-6}$ for all models. We again use the cosine scheduler with a warm-up ratio of $0.1$. The weight $\beta$ and $\alpha$ are set as $0.1$ and $1.0$. LoRA settings are consistent with those in the supervised fine-tuning.

\section{Additional Experiments}
\subsection{Extended Baselines Comparison} \label{sec:appendix-baselines-compare}
Table \ref{tab:overall-performance-fine} presents an extended comparison between LitE-SQL and fine-tuned baseline models with various parameter counts. As shown in the table, several prior approaches rely on significantly larger models—up to 14B or 32B parameters—to achieve competitive performance. In contrast, LitE-SQL attains comparable or superior accuracy with much smaller models. Notably, the 7B LitE-SQL model matches or outperforms larger variants of OmniSQL (14B and 32B) and Reasoning-SQL (14B) on both BIRD and Spider 1.0, despite using far fewer parameters. Overall, these results demonstrate that LitE-SQL achieves strong performance under a markedly reduced model budget, highlighting its superior parameter efficiency relative to existing fine-tuning approaches.

\subsection{Efficiency Analysis} \label{sec:appendix-latency-analysis}
\begin{table}[!t]
\renewcommand{\arraystretch}{1}
    \centering
    \resizebox{\linewidth}{!}{
    \begin{tabular}{lccccc}
        \Xhline{1.0pt}
        \textbf{Method} & \textbf{Avg.(s)} & \textbf{Std.(s)} \\ \Xhline{0.5pt}
        {CHESS \cite{DBLP:journals/corr/abs-2405-16755}} & 83.631 & 43.017 \\
        % \textbf{Schema Retriever}                        & \textbf{0.042} & 0.007 \\
        % \cdashline{1-3}[1pt/1pt]
        {OmniSQL \cite{li2025omnisql}}                   & 76.759 & 1.002 \\
        % \textbf{SQL Generator}                           & \textbf{24.94} & 31.41 \\
        \cdashline{1-3}[1pt/1pt]
        \textbf{LitE-SQL}                           & \textbf{24.979} & 31.414 \\
        \Xhline{1.0pt}
    \end{tabular}
    }
    \caption{Latency on subsampled-BIRD compared to baselines.}
    \label{tab:latency_result}
\end{table}

We evaluate the computational efficiency of LitE-SQL by measuring inference latency. We compare against two representative baselines that follow distinct learning paradigms: CHESS (in-context learning) and OmniSQL (fine-tuning with a 7B-parameter model). For all methods, latency is measured on a per-instance basis, and the reported mean and standard deviation are computed over all instances in subsampled-BIRD.

As shown in Table \ref{tab:latency_result}, both CHESS and OmniSQL incur substantial inference latency, with average per-instance runtimes of 83.631s and 76.759s, respectively. In contrast, LitE-SQL achieves significantly lower latency, reducing the average inference time to 24.979s. This improvement reflects the efficiency of the proposed pipeline, which avoids repeated large-model invocations during schema selection and eliminates the need for generating multiple candidates. Overall, the results demonstrate that LitE-SQL substantially reduces inference-time overhead compared to representative in-context learning and fine-tuning baselines, highlighting its practical efficiency at the system level.

\subsection{Hyperparameter Analysis} \label{sec:appendix-retriever-hyperparameters}
We report supplementary experiments that guided the choice of hyperparameters for the Schema Retriever. We analyze three hyperparameters: margin values for HN-SupCon loss, the number of negative samples for training ($N_i$), and the number of retrieved columns ($k$). \\
\textbf{Margin values.} We investigate the margin hyperparameter in the HN-SupCon loss, which defines the threshold for distinguishing hard negatives. Table~\ref{appendix:hnsupcon_margin} reports results obtained under the configuration of eight negative samples and top-$25$ retrieval columns, consistent with the configurations of $N_i$ and $k$ analyzed below. A margin of 0.1 provides the best trade-off, yielding the highest SLR while maintaining low FPR. Smaller margins (e.g., 0) fail to sufficiently separate hard negatives, whereas larger margins (e.g., 0.2) relax the criterion and reduce recall. Since performance already degrades at 0.2, further increasing the margin is expected to exacerbate this trend; we therefore restrict our analysis to $\{0, 0.1, 0.2\}$ and fix the margin to 0.1 in all experiments. \\
\textbf{Negative sample sizes.} The number of negative samples ($N_i$) affects both training efficiency and schema linking performance. We experimented with values from 3 to 8, with results reported in Tables~\ref{appendix:retriever-tab1-bird} and \ref{appendix:retriever-tab1-spider}. Across both datasets, larger sample sizes generally improved SLR by exposing the model to more challenging negatives, but gains saturated beyond $N_i=5$. In particular, $N_i=8$ consistently achieved the highest SLR under the top-25 setting (87.42 on BIRD and 99.16 on Spider), while maintaining a balanced trade-off with FPR. Adding more negatives would mostly introduce easy cases with limited contribution, while also increasing memory cost. We therefore fix $N_i=8$ in all experiments.\\
\textbf{Retrieved Columns.} The number of retrieved columns provided to the SQL Generator has a direct impact on EX. We empirically evaluate top-$k \in \{10, 15, 20, 25\}$ with SQL Generators of 1.5B, 3B, and 7B parameters. Results are summarized in Table~\ref{appendix:retriever-tab2-total}. On BIRD, top-25 consistently yields the best EX across all model sizes, reflecting the dataset’s large and complex schemas, which require retrieving more columns to ensure coverage of relevant ones. In contrast, Spider contains smaller schemas and most queries involve only 1–4 relevant columns, so retrieving beyond top-15 introduces additional noise without substantial benefit. This effect is most evident for smaller models (1.5B and 3B), which are less robust to irrelevant columns. The 7B model, however, is better able to disregard noise and thus achieves peak performance at top-25 on both datasets. Accordingly, we set $k=25$ in our final configuration.

\section{Input Templates} \label{sec:full prompts}
\subsection{SQL Generator}
For all datasets, we use the same input template for prompt design as illustrated in Table~\ref{tab:input_template_sft} and Table~\ref{tab:input_template_rft}.
\subsection{Schema Retriever}
We provide the input templates—query and schema-document formats—used for training the embedding model (Tables~\ref{tab:query-format} and \ref{tab:doc-format}).

\begin{table*}[t]
    \centering
    \resizebox{1\linewidth}{!}{
    \begin{tabular}{cccccccccc}
    \Xhline{1.0pt}
    \multirow{2}{*}{\textbf{Models}} & \multicolumn{3}{c}{\textbf{NegLimit\_3}} & \multicolumn{3}{c}{\textbf{NegLimit\_4}} & \multicolumn{3}{c}{\textbf{NegLimit\_5}} \\ \cline{2-10} 
    & \textbf{TPR (↑)} & \textbf{FPR (↓)} & \textbf{SLR (↑)} & \textbf{TPR (↑)} & \textbf{FPR (↓)} & \textbf{SLR (↑)} & \textbf{TPR (↑)} & \textbf{FPR (↓)} & \textbf{SLR (↑)} \\
    \Xhline{0.5pt}
    Top10 & 86.19 & 62.36 & 60.43 & 87.05 & 61.97 & 62.52 & 87.51 & 61.58 & 61.34 \\
    Top15 & 91.48 & 72.12 & 71.25 & 92.13 & 71.92 & 74.32 & 92.83 & 71.66 & 75.68 \\
    Top20 & 94.34 & 77.49 & 78.75 & 94.54 & 77.48 & 80.77 & 95.32 & 77.26 & 82.79 \\
    Top25 & 95.90 & 80.87 & 83.64 & 96.04 & 80.85 & 85.14 & 96.69 & 80.72 & 87.29 \\ 
    \Xhline{1.0pt}
    \multirow{2}{*}{\textbf{Models}} & \multicolumn{3}{c}{\textbf{NegLimit\_6}} & \multicolumn{3}{c}{\textbf{NegLimit\_7}} & \multicolumn{3}{c}{\textbf{NegLimit\_8}} \\ \cline{2-10} 
    & \textbf{TPR (↑)} & \textbf{FPR (↓)} & \textbf{SLR (↑)} & \textbf{TPR (↑)} & \textbf{FPR (↓)} & \textbf{SLR (↑)} & \textbf{TPR (↑)} & \textbf{FPR (↓)} & \textbf{SLR (↑)} \\ 
    \Xhline{0.5pt}
    Top10 & 87.96 & 61.49 & 63.36 & 87.52 & 61.74 & 63.10 & 89.14 & 60.85 & \textbf{66.69} \\
    Top15 & 92.67 & 71.75 & 75.68 & 92.64 & 71.78 & 74.84 & 93.49 & 71.45 & \textbf{78.36} \\
    Top20 & 94.90 & 77.37 & 82.07 & 95.24 & 77.31 & 82.27 & 95.54 & 77.20 & \textbf{83.83} \\
    Top25 & 96.05 & 80.85 & 85.40 & 96.78 & 80.71 & 86.70 & 96.71 & 80.71 & \underline{\textbf{87.42}} \\ 
    \Xhline{1.0pt}
    \end{tabular}
    }
    \caption{Comparison of different numbers of negative samples and retrieved columns on BIRD development set.}
    \label{appendix:retriever-tab1-bird}
\end{table*}

\begin{table*}[t]
    \centering
    \resizebox{1\linewidth}{!}{
    \begin{tabular}{cccccccccc}
    \Xhline{1.0pt}
    \multirow{2}{*}{\textbf{Models}} & \multicolumn{3}{c}{\textbf{NegLimit\_3}} & \multicolumn{3}{c}{\textbf{NegLimit\_4}} & \multicolumn{3}{c}{\textbf{NegLimit\_5}} \\ \cline{2-10} 
    & \textbf{TPR (↑)} & \textbf{FPR (↓)} & \textbf{SLR (↑)} & \textbf{TPR (↑)} & \textbf{FPR (↓)} & \textbf{SLR (↑)} & \textbf{TPR (↑)} & \textbf{FPR (↓)} & \textbf{SLR (↑)} \\
    \Xhline{0.5pt}
    Top10 & 95.00 & 71.38 & 90.50 & 94.86 & 71.42 & 89.85 & 95.06 & 71.34 & 90.03 \\
    Top15 & 96.77 & 79.28 & 95.62 & 96.76 & 79.32 & 95.16 & 96.81 & 79.29 & 95.58 \\
    Top20 & 97.70 & 82.65 & 97.44 & 97.82 & 82.65 & 97.21 & 97.87 & 82.63 & 97.58 \\
    Top25 & 98.67 & 83.99 & 98.46 & 98.91 & 83.98 & 98.70 & 98.95 & 83.97 & 98.88 \\ 
    \Xhline{1.0pt}
    \multirow{2}{*}{\textbf{Models}} & \multicolumn{3}{c}{\textbf{NegLimit\_6}} & \multicolumn{3}{c}{\textbf{NegLimit\_7}} & \multicolumn{3}{c}{\textbf{NegLimit\_8}} \\ \cline{2-10} 
    & \textbf{TPR (↑)} & \textbf{FPR (↓)} & \textbf{SLR (↑)} & \textbf{TPR (↑)} & \textbf{FPR (↓)} & \textbf{SLR (↑)} & \textbf{TPR (↑)} & \textbf{FPR (↓)} & \textbf{SLR (↑)} \\ 
    \Xhline{0.5pt}
    Top10 & 94.93 & 71.43 & 90.13 & 94.74 & 71.52 & 89.01 & 95.15 & 71.34 & \textbf{90.36} \\
    Top15 & 96.87 & 79.31 & 95.34 & 96.70 & 79.31 & 95.34 & 97.11 & 79.28 & \textbf{95.48} \\
    Top20 & 97.82 & 82.66 & 97.25 & 97.84 & 82.64 & 97.39 & 98.20 & 82.62 & \textbf{97.76} \\
    Top25 & 98.82 & 83.99 & 98.65 & 98.89 & 83.98 & 98.79 & 99.27 & 83.96 & \underline{\textbf{99.16}} \\ 
    \Xhline{1.0pt}
    \end{tabular}
    }
    \caption{Comparison of different numbers of negative samples and retrieved columns on Spider 1.0 test set.}
    \label{appendix:retriever-tab1-spider}
\end{table*}

\begin{table*}[t]
    \centering
    \resizebox{1\linewidth}{!}{%
        \begin{tabular}{ccccccccccc}
        \Xhline{1.0pt}
        \textbf{Sizes} & \multicolumn{5}{c}{\textbf{BIRD (Dev)}} & \multicolumn{5}{c}{\textbf{Spider 1.0 (Test)}} \\ 
        \Xhline{0.5pt}
        \textbf{1.5B} & \textbf{Simple} & \textbf{Moderate} & \textbf{Chall.} & \textbf{Overall} & & \textbf{Easy} & \textbf{Medium} & \textbf{Hard} & \textbf{Extra} & \textbf{Overall} \\
        \Xhline{0.5pt}
        Best  & 73.51          & 54.31          & 46.90          & 65.19          & & 92.13          & 85.88          & 77.54          & 71.15          & 83.00 \\
        Top10 & 56.00          & 35.99          & 31.72          & 47.65          & & 88.72          & 79.23          & 68.68          & 64.43          & 76.57 \\
        Top15 & 58.38          & 39.01          & 31.72          & 50.00          & & 88.30          & \textbf{80.05} & \textbf{69.96} & \textbf{65.55} & \textbf{77.27} \\
        Top20 & 58.16          & 39.87          & 32.41          & 50.20          & & \textbf{89.36} & 79.46          & 69.55          & 63.87          & 76.90 \\
        Top25 & \textbf{59.46} & \textbf{40.30} & \textbf{35.86} & \textbf{51.43} & & 89.15          & 78.76          & 69.11          & 64.99          & 76.67\\ 
        \Xhline{1.0pt}
        \textbf{3B} & \textbf{Simple} & \textbf{Moderate} & \textbf{Chall.} & \textbf{Overall} & & \textbf{Easy} & \textbf{Medium} & \textbf{Hard} & \textbf{Extra} & \textbf{Overall} \\ 
        \Xhline{0.5pt}
        Best  & 76.32          & 63.15          & 51.72          & 70.01          & & 94.89          & 88.10          & 78.83          & 74.23          & 85.28 \\
        Top10 & 58.16          & 46.77          & 37.24          & 52.74          & & 90.21          & \textbf{84.36} & 74.73          & 64.99          & 80.34 \\
        Top15 & 59.78          & 49.78          & 43.45          & 55.22          & & \textbf{90.21} & 84.25          & \textbf{77.75} & \textbf{65.27} & \textbf{81.00} \\
        Top20 & \textbf{62.05} & 51.51          & 40.69          & 56.84          & & 89.79          & 84.01          & 74.95          & 62.75          & 79.79 \\
        Top25 & 61.95          & \textbf{51.29} & \textbf{43.45} & \textbf{56.98} & & 89.15          & 83.20          & 74.08          & 61.90          & 78.99 \\ 
        \Xhline{1.0pt}
        \textbf{7B} & \textbf{Simple} & \textbf{Moderate} & \textbf{Chall.} & \textbf{Overall} & & \textbf{Easy} & \textbf{Medium} & \textbf{Hard} & \textbf{Extra} & \textbf{Overall} \\ 
        \Xhline{0.5pt}
        Best  & 78.59          & 63.58          & 57.93          & 72.10          & & 95.74          & 91.60          & 82.51          & 78.99          & 88.45 \\
        Top10 & 62.38          & 42.46          & 40.00          & 54.24          & & \textbf{92.77} & 85.76          & 77.75          & 70.31          & 83.00 \\
        Top15 & 64.86          & 48.92          & 43.45          & 58.02          & & 92.55          & 86.00          & 80.13          & 72.27          & 83.88 \\
        Top20 & 66.59          & 50.00          & 47.59          & 59.78          & & 92.34          & 86.35          & 79.70          & 72.55          & 83.93 \\
        Top25 & \textbf{67.78} & \textbf{50.86} & \textbf{45.52} & \textbf{60.56} & & 92.13          & \textbf{86.46} & \textbf{81.21} & \textbf{73.11} & \textbf{84.35} \\ 
        \Xhline{1.0pt}
        \end{tabular}
    }
    \caption{Effect of the number of retrieved columns (top-$k$) on EX with SQL Generators of 1.5B, 3B, and 7B parameters. Results are reported by question difficulty (BIRD: simple, moderate, Chall.; Spider: easy, medium, hard, extra) as well as Overall performance. Chall.\ denotes challenging, and Overall denotes the aggregate performance across all difficulty levels.}
    \label{appendix:retriever-tab2-total}
\end{table*}

\onecolumn
\begin{tcolorbox}[colback=gray!10!white, colframe=black, title=\textbf{\textsc{Input Template of SFT}}, width=\linewidth, boxrule=1pt]
\#\#\# Question \newline
\{QUESTION\} \newline

\#\#\# Database \newline
-- Tables and Columns \newline
\{COLUMN\_SCHEMA\_INFORMATION\}
\begin{verbatim}
# Example:
disp.disp_id = {
  "type": "integer",
  "primary_key": true,
  "values": [9, 2],
  "description": "unique number of identifying this row of record",
  "comment": "disposition id"
}
...
\end{verbatim}

-- Foreign Keys \newline
\{FOREIGN\_KEYS\_RELATIONS\}
\begin{verbatim}
# Example:
disp.disp_id = card.disp_id
...
\end{verbatim}
-- Evidence \newline
\{EVIDENCE\}
\end{tcolorbox}
\begin{minipage}{\textwidth}
\captionof{table}{Input template of SFT stage with example for column schema and foreign keys.}
\label{tab:input_template_sft}
\end{minipage}

\begin{tcolorbox}[colback=gray!10!white, colframe=black, title=\textbf{\textsc{Input Template of RFT}}, width=\linewidth, boxrule=1pt]
\{INPUT TEMPLATE OF SFT\} \newline

\{SQL GENERATED WRONG\} \newline

\#\#\# Error Message \newline
\{ERROR MESSAGE\}

\end{tcolorbox}
\begin{minipage}{\textwidth}
\captionof{table}{Input template of RFT stage with generated wrong SQL and corresponding execution feedback.}
\label{tab:input_template_rft}
\end{minipage}

\begin{tcolorbox}[colback=gray!10!white, colframe=black, title=\textbf{\textsc{Input Template of Query}}, width=\linewidth, boxrule=1pt]
Instruct:Given a natural language question, retrieve database column information passages used to generate SQL.\newline
Query:\{QUESTION\} \{EVIDENCE\}
\end{tcolorbox}
\begin{minipage}{\textwidth}
\captionof{table}{Input format used for training the embedding model. The \{EVIDENCE\} field is optional and omitted when unavailable (e.g., Spider).}
\label{tab:query-format}
\end{minipage}

\begin{tcolorbox}[colback=gray!10!white, colframe=black, title=\textbf{\textsc{Input Template of Document}}, width=\linewidth, boxrule=1pt]
table:\{TABLE\_NAME\} \newline
column:\{COLUMN\_NAME\} \newline
column\_desc:\{COLUMN\_DESCRIPTION\} \newline
value\_desc:\{VALUE\_DESCRIPTION\}
\end{tcolorbox}
\begin{minipage}{\textwidth}
\captionof{table}{Input format for database schema documents used in training the embedding model and stored in the vector database. The column\_desc and value\_desc fields are optional and omitted when not available.}
\label{tab:doc-format}
\end{minipage}

\end{document}